\documentclass[10pt,twocolumn,letterpaper]{article}

\usepackage{wacv}              

\usepackage{float}

\definecolor{wacvblue}{rgb}{0.21,0.49,0.74}
\usepackage[pagebackref,breaklinks,colorlinks,allcolors=wacvblue]{hyperref}

\def\wacvPaperID{*****} 
\def\confName{WACV}
\def\confYear{2027}

\title{LegSegNet: A Public Deep Learning System for Lower Extremity CT Tissue Segmentation and Quantification}

\author{
\begin{tabular}{c}
Yuwen Chen$^{1}$ \quad
Yaqian Chen$^{1}$ \quad
Roy Colglazier$^{3}$ \quad
Haoyu Dong$^{1}$\\
Hanxue Gu$^{1}$ \quad
Maciej A. Mazurowski$^{1,2,3,4}$ \quad
Kevin W. Southerland$^{5}$\\
{\small $^{1}$Department of Electrical and Computer Engineering, Duke University, Durham, NC 27708}\\
{\small $^{2}$Department of Biostatistics \& Bioinformatics, Duke University, Durham, NC 27708}\\
{\small $^{3}$Department of Radiology, Duke University, Durham, NC 27708}\\
{\small $^{4}$Department of Computer Science, Duke University, Durham, NC 27708}\\
{\small $^{5}$Department of Surgery, Duke University, Durham, NC 27708}
\end{tabular}
}

\begin{document}
\maketitle

\begin{abstract}
\label{sec:abstract}
Lower extremity computed tomography (CT) contains clinically relevant information for body composition analysis, sarcopenia assessment, and musculoskeletal disease monitoring, but extracting these measurements at scale requires accurate tissue segmentation and an automated quantification workflow. Existing public segmentation tools are not designed for comprehensive lower extremity CT analysis, particularly for clinically important inter/intramuscular adipose tissue, and most public methods only provide mask prediction rather than an end-to-end quantification system. To address this problem, we present LegSegNet, a deep learning system for lower extremity CT tissue segmentation and body composition quantification. Given an input CT scan, LegSegNet segments bone, skeletal muscle, subcutaneous adipose tissue, and inter/intramuscular adipose tissue. It then computes quantitative tissue measurements for downstream analysis. We developed the segmentation model using 1,302 manually annotated CT slices and evaluated it on 900 held-out test slices, with all annotations reviewed by radiologists. We benchmark LegSegNet against a broad set of 2D segmentation methods, including CNN-based models, transformer-based models, and finetuned foundation models, and further evaluate its generalization on an external public CT dataset. LegSegNet achieves the best overall segmentation performance, with an average Dice score of 89.31 on the held-out test set. To our knowledge, LegSegNet is the first publicly available end-to-end system for lower extremity CT tissue segmentation and quantification, providing a practical evaluation tool for future computer vision research in medical image analysis. The code and model weights are available at: \href{https://github.com/mazurowski-lab/LegSegNet}{https://github.com/mazurowski-lab/LegSegNet}

\end{abstract}

\begin{table*}[t]
\centering
\caption{Comparison of CT lower extremity deep learning segmentation models and systems for body composition analysis.}
\label{tab:system}
\footnotesize
\setlength{\tabcolsep}{3pt}
\renewcommand{\arraystretch}{1.12}
\begin{tabular}
{p{0.19\textwidth} | p{0.34\textwidth} | p{0.28\textwidth} | p{0.11\textwidth}}
\toprule
\textbf{Models / Papers} & \textbf{Labels} & \textbf{Body Range} & \textbf{Public Weights} \\
\midrule
Yang \etal, 2022 \cite{yang2022quantification} & Muscle, bone, subcutaneous adipose tissue, intermuscular adipose tissue & Thigh & Not reported \\
\midrule
Yoo \etal, 2022 \cite{yoo2022deep} & Three thigh muscle groups and adipose tissue & Thigh & Not reported \\
\midrule
TotalSegmentator, 2023 \cite{wasserthal2023totalsegmentator} & Broad CT anatomy, including part of lower extremity bones and selected muscle labels in pelvis-to-knee region & Whole-body & Yes \\
\midrule
Soufi \etal, 2025 \cite{soufi2025validation} & 19 hip and thigh muscles and 3 bones & Hip-to-knee & Not reported \\
\midrule
Imani \etal, 2025 \cite{imani2025deep} & Bone, bone marrow adipose tissue, skeletal muscle, intermuscular adipose tissue, subcutaneous adipose tissue & Hip & Not reported \\
\midrule
Na \etal, 2025 \cite{na2025automated} & Fat, muscle, and fluid-fibrotic tissue & Lower extremity & Not reported \\
\midrule
Kim \etal, 2025 \cite{kim2025deep} & Thigh muscle & Thigh & Not reported \\
\midrule
LegSegNet (ours) & Bone, skeletal muscle, subcutaneous adipose tissue, inter/intramuscular adipose tissue & Lower extremity & Yes \\
\bottomrule
\end{tabular}
\end{table*}

\section{Introduction}
\label{sec:intro}

CT-based body composition analysis is widely used for assessing musculoskeletal health, metabolic disorders, and disease prognosis \cite{prado2014lean,melba:2025:026:chen}. In particular, quantitative assessment of lower extremity tissue volume and spatial distribution, including bone, skeletal muscle, and adipose tissue, can provide valuable imaging biomarkers for diseases such as sarcopenia, osteoporosis, obesity, and peripheral vascular disease \cite{cruz2019sarcopenia,tolonen2021methodology,engelke2018quantitative}. However, reliable and efficient extraction of these biomarkers requires accurate tissue segmentation.

Manual segmentation and annotation of these tissues are labor-intensive and require domain expertise, which limits the scalability of body composition studies. To reduce this annotation burden, prior studies have explored a range of strategies for maintaining model robustness under limited annotation, including increasing the number of annotated volumes under a fixed annotation budget \cite{zhang2024select}, leveraging domain adaptation across different imaging modalities \cite{melba:2025:031:chen}, and using self-supervised or semi-supervised learning with unlabeled data \cite{chen2023moco,dong2025mri}.

Beyond methods development, public datasets and released model weights have also played an important role in enabling reproducible evaluation, external validation, and broader adoption of segmentation tools \cite{wasserthal2023totalsegmentator,gu2025segmentanybone,colglazier2025segmentanymuscle}. However, existing public tools remain limited for comprehensive lower extremity CT tissue segmentation.

For example, TotalSegmentator \cite{wasserthal2023totalsegmentator} provides related masks such as bone and thigh-based muscle, but these tasks do not constitute a full lower extremity tissue model. Existing public CT body composition models primarily focus on the chest, abdomen, and pelvis rather than the lower extremities \cite{melba:2025:026:chen}. Prior lower extremity CT studies have explored lymphedema tissue segmentation \cite{na2025automated}, thigh muscle and bone assessment \cite{soufi2025validation}, and thigh muscle segmentation \cite{kim2025deep}. However, these models are often disease-specific, anatomically restricted, or not released for publicly available use. These limitations highlight the demand for a publicly available model for comprehensive lower extremity CT tissue segmentation.

To address this issue, we present \textbf{LegSegNet}, a deep learning framework specifically designed for multi-compartment segmentation of lower extremity CT scans, with public model weights for unified segmentation of bone, skeletal muscle, subcutaneous adipose tissue, and inter/intramuscular adipose tissue. We also develop a publicly available application for quantitative lower extremity CT analysis. Our main contributions are summarized below:
\begin{itemize}
    \item We provide a multi-class lower extremity CT segmentation model for bone, muscle, and adipose tissue.
    \item We develop a comprehensive benchmark including multiple segmentation architectures, including CNN-based and transformer-based methods, and demonstrate that LegSegNet achieves the best average performance.
    \item We introduce LegSegNet, which, to our knowledge, is the first publicly available end-to-end system for lower extremity CT tissue quantification and analysis, and we have made the code and model weights publicly available.
\end{itemize}

\section{Related Work}
\label{sec:related_work}

\paragraph{Computer vision for medical image segmentation.}
Computer vision methods have been widely used for medical image segmentation, enabling automated extraction of anatomical structures and quantitative biomarkers from large imaging cohorts. Convolutional neural networks (CNNs), such as UNet \cite{ronneberger2015u} and its variants \cite{oktay2018attention,myronenko20183d,tan2019rethinking}, are still commonly used for biomedical segmentation. More recently, transformer-based models and foundation segmentation models, such as SAM \cite{kirillov2023segment} and MedSAM \cite{ma2024segment}, have also been explored and developed for medical imaging tasks. Among these methods, nnUNet remains one of the state-of-the-art (SOTA) baselines \cite{isensee2021nnu,isensee2024nnu}, as it automatically adapts preprocessing, training pipeline, and post-processing to the target dataset. Therefore, in this work, we use nnUNet as the basis for LegSegNet development and benchmark representative CNN-based and transformer-based methods.

\paragraph{CT body composition analysis.}
CT-derived measurements of skeletal muscle, adipose tissue, and bone can provide clinically useful biomarkers for nutritional assessment, sarcopenia, osteoporosis, obesity, and disease prognosis \cite{prado2014lean,cruz2019sarcopenia,tolonen2021methodology,engelke2018quantitative}. Traditional body composition analysis often relies on manual or semi-automated delineation of selected axial slices or anatomical regions, which limits efficiency and introduces observer variability. Thus, deep learning has become an important approach for scalable tissue quantification. Recent CT models have automated muscle and fat segmentation across the chest, abdomen, and pelvis \cite{melba:2025:026:chen,pu2023automated}. However, these systems are not designed to provide a comprehensive lower extremity tissue quantification covering bone, skeletal muscle and adipose tissues. This motivates the development of an automated lower extremity CT tissue segmentation and quantification system.

\paragraph{Public segmentation models for CT lower extremity segmentation.}
Public systems specifically designed for lower extremity CT tissue segmentation and quantification remain limited. TotalSegmentator provides a broad, general-purpose CT segmentation system for many anatomical structures and includes part of the lower extremity labels such as bones and selected muscle groups \cite{wasserthal2023totalsegmentator}. However, it mainly focuses on thigh-based muscle and is not designed as an end-to-end lower extremity body composition system for joint quantification of bone, skeletal muscle and adipose tissue. In addition, other lower extremity CT studies have focused on narrower segmentation settings, including lymphedema tissue segmentation \cite{na2025automated}, hip-to-knee bone and muscle assessment \cite{soufi2025validation}, and thigh muscle segmentation \cite{kim2025deep}. These studies show the feasibility of deep learning for lower extremity CT analysis, but existing models are commonly disease-specific, anatomically restricted, or unavailable for public use (As shown in Table \ref{tab:system}). Moreover, most existing models do not provide a unified quantification of clinically important inter/intramuscular adipose tissue \cite{yim2007intermuscular,zhang2023intermuscular}. Thus, LegSegNet aims to close this gap by focusing on comprehensive lower extremity CT tissue segmentation and quantification, with a public release of the system and model weights.

\section{Methods}
\label{sec:methods}

\subsection{Cohort Selection}

We selected 107 patients for model development and an additional 30 patients for independent model evaluation who took CT examination at the Duke Health System between 2013 and 2020. The development cohort mainly consists of two CT protocols: CT chest abdomen pelvis with contrast and CT angiography with runoff. All imaging data were deidentified before export.

\begin{figure}[t]
\centering
\includegraphics[width=\linewidth]{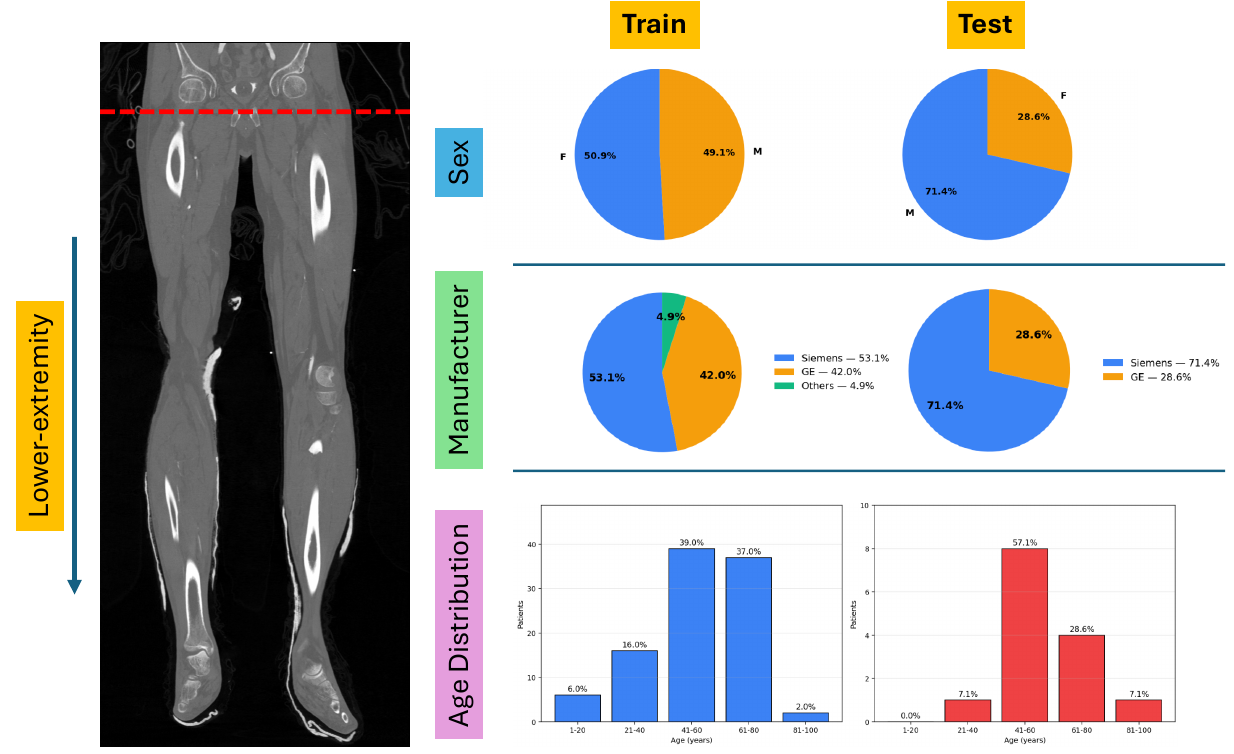}
\caption{\textbf{Dataset overview.} Left: Example coronal CT view of the lower extremities with the annotated region indicated. Right: Demographics of the training and test sets, including sex distribution, CT scanner manufacturer, and age distribution.}
\label{fig:cohort}
\end{figure}

For both protocols, we only annotate axial slices below the pelvis, as shown in Figure \ref{fig:cohort}. Based on \cite{zhang2024select}, instead of annotating the whole volume, we randomly annotate between 1 and 50 axial slices (median 3) for each volume in the development cohort. This results in 1,302 annotated 2D slices for model development, which are then split by patient into 1,041 training slices and 261 validation slices. For the independent test set, we select 30 axial slices evenly below the pelvis from each of the 30 held-out patients, resulting in 900 testing slices.

We further test the model's generalizability on external test set. We randomly select 30 whole-body CT scans from \textbf{SAROS}~\cite{koitka2024saros}, a publicly available multi-center dataset, and evaluate LegSegNet on the lower extremity range. SAROS provides annotations for subcutaneous adipose tissue and skeletal muscle.

\subsection{Label Description}

Each axial CT slice has been annotated into four foreground classes. Subcutaneous adipose tissue (\textbf{SAT}) is defined as the soft tissue between the skin and the deep muscular fascia, and therefore includes the subcutaneous fat layer as well as related soft-tissue changes within this region. Skeletal muscle (\textbf{SM}) includes the visible muscle of the thigh and calf. Inter/intramuscular adipose tissue (\textbf{IAT}) refers to fat located within the muscle, either between muscle groups or within muscle tissue. \textbf{Bone} includes the visible bone structures, including femur, tibia, fibula, and patella. Small structures that are not part of the target label set, such as small vessels and edema, are assigned to the surrounding label to maintain a consistent compartment-level annotation strategy, similar as \cite{pu2023automated}. Therefore, these labels provide a practical representation of major lower extremity tissues for CT-based body composition analysis.

\subsection{Annotation Pipeline}

We adopt an interactive model-in-the-loop annotation pipeline to improve annotation efficiency. To do this, we first randomly select 10 volumes to annotate from scratch. A 2D nnUNet is then trained using the available annotations and used to generate pseudo masks for the next batch of volumes. The annotators load these pseudo masks into 3D Slicer \cite{fedorov20123d} and manually correct the sampled slices. The corrected masks are then added back to the training set, and the model is retrained for the next annotation round.

\subsection{Segmentation Methods}

LegSegNet is built upon nnUNet \cite{isensee2021nnu}, a self-configuring deep learning framework that has become one of the strongest and most widely used approaches for biomedical image segmentation \cite{isensee2024nnu,li2025breastsegnet}. Unlike manually designed pipelines, nnUNet automatically configures key components of the segmentation workflow, including preprocessing, patch size, network architecture, training pipelines, data augmentation, and postprocessing, based on an extracted dataset-specific fingerprint. In this study, we evaluate the default 2D nnUNet configuration and two residual-encoder variants, nnUNet 2D ResEnc-L and nnUNet 2D ResEnc-XL (See Table \ref{tab:segmentation_models} for details).

\begin{figure*}[t]
\centering
\includegraphics[width=\textwidth]{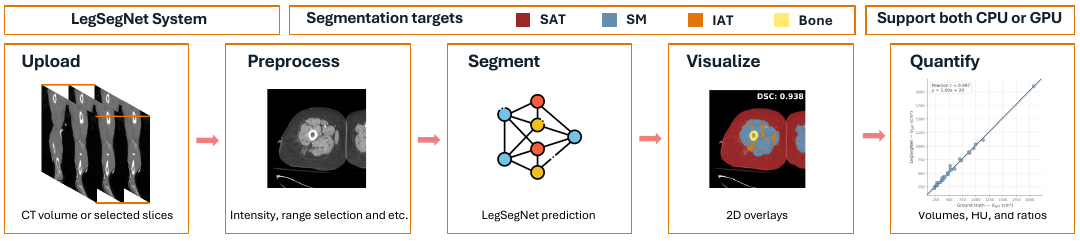}
\caption{\textbf{Overall pipeline of LegSegNet.} The system takes lower extremity CT scans as input, performs automated tissue segmentation, and generates visualizations and quantitative body composition measurements for downstream analysis.}
\label{fig:pipeline}
\end{figure*}

\subsection{Automated Quantification System Design}

In addition to mask prediction, LegSegNet is designed as an automated lower extremity CT tissue quantification system. As shown in Figure \ref{fig:pipeline}, we implement the system with Gradio interface, which allows users to upload CT volumes or slices, run the trained segmentation model, visualize the predicted masks, and export quantitative measurements. The system supports both CPU and GPU inference, allowing it to run on standard local computers while also taking advantage of GPU acceleration when available. The system applies the trained model at slice-level, reconstructs the predicted labels in the original image space, and computes tissue measurements from the resulting masks.

For each tissue class $c \in \{\textbf{SAT}, \textbf{SM}, \textbf{IAT}, \textbf{Bone}\}$ and axial slice $k$, the cross-sectional area can be computed as following:

\begin{equation}
A_{c,k} = s_x \times s_y \times N_{c,k}
\end{equation}

where $N_{c,k}$ is the number of pixels assigned to class $c$ and $s_x, s_y$ are the pixel spacings. Then, tissue volume can be computed by summing slice areas across the analyzed lower extremity region:

\begin{equation}
V_c = \sum_k A_{c,k} \cdot \Delta z_k
\end{equation}

where $\Delta z_k$ is the slice thickness. The system also computes mean CT attenuation for each class as below:

\begin{equation}
\mu_c = \frac{1}{N_c}\sum_{i \in M_c} I_i
\end{equation}

where $M_c$ is the predicted mask for class $c$, $N_c$ is the number of voxels in the mask, and $I_i$ is the CT intensity in Hounsfield units.

Based on the available labels, LegSegNet can report a set of directly computable lower extremity body composition metrics. Total adipose tissue volume is defined as the sum of subcutaneous and inter/intramuscular adipose tissue as:

\begin{equation}
V_{\mathrm{Fat}} = V_{\mathbf{SAT}} + V_{\mathbf{IAT}}.
\end{equation}

The IAT-to-SAT ratio computes relative fat infiltration within the muscle compared with subcutaneous fat as:

\begin{equation}
R_{\mathbf{IAT}/\mathbf{SAT}} = \frac{V_{\mathbf{IAT}}}{V_{\mathbf{SAT}} + \epsilon}
\end{equation}

where $\epsilon=10^{-9}$ is a small positive constant added only to the denominator to avoid division by zero when the denominator tissue volume is zero or extremely small. The muscle-to-fat ratio is computed as:

\begin{equation}
R_{\mathbf{SM}/\mathrm{Fat}} = \frac{V_{\mathbf{SM}}}{V_{\mathbf{SAT}} + V_{\mathbf{IAT}} + \epsilon}
\end{equation}

In addition, the system presents the skeletal muscle attenuation and adipose tissue attenuation as below:

\begin{equation}
\mu_{\mathbf{SM}} = \frac{1}{N_{\mathbf{SM}}}\sum_{i \in M_{\mathbf{SM}}} I_i
\end{equation}
\begin{equation}
\mu_{\mathrm{Fat}} = \frac{1}{N_{\mathbf{SAT}} + N_{\mathbf{IAT}}}\left(\sum_{i \in M_{\mathbf{SAT}}} I_i + \sum_{i \in M_{\mathbf{IAT}}} I_i\right)
\end{equation}

Moreover, LegSegNet can summarize area-based measurements at each axial position to produce longitudinal tissue profiles $\{A_{c,k}\}_{k=1}^{K}$ along the lower extremity, which can be useful for identifying regional muscle loss or abnormal fat distribution.

\section{Experiments}
\label{sec:experiments}

\begin{table*}[t]
\centering
\caption{Model performance on held-out test set including per-class 3D Dice ($\uparrow$) and ASSD ($\downarrow$) with 95\% bootstrap confidence intervals. The best is bold and runner-up is underlined.}
\label{tab:test_results_3d}
\resizebox{1\textwidth}{!}{
\begin{tabular}{l|cccc|c|cccc|c}
\toprule
& \multicolumn{5}{c|}{Dice $\uparrow$} & \multicolumn{5}{c}{ASSD $\downarrow$} \\
\midrule
Method & \textbf{SAT} & \textbf{SM} & \textbf{IAT} & \textbf{Bone} & Ave. & \textbf{SAT} & \textbf{SM} & \textbf{IAT} & \textbf{Bone} & Ave. \\
\midrule
\multicolumn{11}{l}{\textit{CNN-based}} \\
  UNet & 92.11 [90.56, 93.39] & 91.51 [89.72, 93.12] & 70.92 [68.21, 73.46] & 96.04 [95.34, 96.59] & 87.65 [86.40, 88.83] & 0.356 [0.295, 0.423] & 0.593 [0.460, 0.754] & 1.191 [1.011, 1.383] & 0.808 [0.277, 1.560] & 0.737 [0.570, 0.947] \\
  Attention UNet & 92.46 [90.94, 93.72] & 92.10 [90.39, 93.64] & 70.08 [66.70, 73.27] & 96.09 [94.77, 96.98] & 87.68 [86.30, 89.06] & 0.474 [0.300, 0.768] & 0.517 [0.412, 0.643] & 1.295 [1.071, 1.529] & 0.930 [0.234, 2.174] & 0.804 [0.565, 1.115] \\
  SegResNet & 92.61 [91.12, 93.89] & 92.63 [91.08, 94.06] & 72.19 [69.64, 74.68] & 96.80 [96.26, 97.20] & 88.56 [87.43, 89.65] & 0.435 [0.303, 0.654] & 0.445 [0.364, 0.532] & 1.146 [0.947, 1.376] & 0.623 [0.192, 1.366] & 0.662 [0.502, 0.855] \\
  EffNet-B0 UNet & 92.04 [90.55, 93.26] & 91.63 [89.90, 93.25] & 69.66 [66.95, 72.25] & 96.16 [95.60, 96.68] & 87.37 [86.13, 88.55] & 0.364 [0.302, 0.426] & 0.532 [0.434, 0.659] & 1.266 [1.081, 1.476] & 0.687 [0.283, 1.398] & 0.712 [0.569, 0.895] \\
\midrule
\multicolumn{11}{l}{\textit{Transformer-based}} \\
  SwinUNETR & 92.16 [90.78, 93.31] & 91.43 [89.77, 93.00] & 69.08 [66.47, 71.50] & 95.18 [93.82, 96.16] & 86.96 [85.72, 88.17] & 0.363 [0.297, 0.434] & 0.559 [0.448, 0.701] & 1.263 [1.090, 1.457] & 0.944 [0.312, 2.132] & 0.782 [0.571, 1.088] \\
  SAM ViT-B & 90.93 [89.55, 92.10] & 90.78 [89.19, 92.28] & 68.55 [66.21, 70.79] & 95.35 [94.86, 95.81] & 86.40 [85.30, 87.49] & 0.399 [0.342, 0.458] & 0.592 [0.492, 0.704] & 1.175 [1.014, 1.362] & 0.483 [0.249, 0.878] & 0.662 [0.564, 0.779] \\
  MedSAM ViT-B & 82.48 [80.39, 84.26] & 88.21 [86.18, 90.12] & 55.02 [51.97, 57.98] & 90.96 [89.78, 92.03] & 79.17 [77.80, 80.54] & 0.774 [0.660, 0.896] & 0.852 [0.687, 1.062] & 1.792 [1.564, 2.053] & 0.724 [0.428, 1.235] & 1.036 [0.883, 1.225] \\
\midrule
\multicolumn{11}{l}{\textit{LegSegNet}} \\
  nnUNet (default 2D) & 93.19 [91.83, 94.36] & \textbf{92.92 [91.43, 94.24]} & \textbf{73.98 [71.27, 76.54]} & \textbf{97.15 [96.82, 97.43]} & \textbf{89.31 [88.18, 90.41]} & 0.299 [0.246, 0.354] & \textbf{0.404 [0.340, 0.471]} & \textbf{1.062 [0.863, 1.291]} & \textbf{0.296 [0.130, 0.572]} & \textbf{0.515 [0.434, 0.598]} \\
  nnUNet ResEnc-L & \textbf{93.27 [91.94, 94.40]} & 92.76 [91.26, 94.13] & 72.42 [69.55, 75.05] & 97.05 [96.68, 97.36] & 88.87 [87.67, 90.01] & \underline{0.290 [0.241, 0.338]} & 0.418 [0.353, 0.487] & 1.138 [0.913, 1.403] & \underline{0.331 [0.133, 0.680]} & \underline{0.544 [0.452, 0.640]} \\
  nnUNet ResEnc-XL & \underline{93.23 [91.83, 94.42]} & \underline{92.79 [91.26, 94.15]} & \underline{73.09 [70.44, 75.64]} & \underline{97.10 [96.75, 97.38]} & \underline{89.05 [87.88, 90.14]} & \textbf{0.288 [0.240, 0.337]} & \underline{0.406 [0.340, 0.474]} & \underline{1.133 [0.889, 1.452]} & 0.503 [0.155, 1.001] & 0.583 [0.455, 0.727] \\
\bottomrule
\end{tabular}
}
\end{table*}

\subsection{Baselines}
\begin{table}[t]
\centering
\caption{Overview of evaluated segmentation models in benchmark.}
\label{tab:segmentation_models}
\resizebox{\linewidth}{!}{%
\begin{tabular}{l|c|c|c|c}
\toprule
\textbf{Model} & \textbf{Family} & \textbf{Architecture} & \textbf{Pretraining} & \textbf{Training} \\
\midrule
nnUNet 2D (plain) & nnUNet & PlainConvUNet & -- & full \\
nnUNet 2D ResEnc-L & nnUNet & Residual-encoder UNet (L) & -- & full \\
nnUNet 2D ResEnc-XL & nnUNet & Residual-encoder UNet (XL) & -- & full \\
UNet & CNN & Residual UNet & -- & full \\
Attention UNet & CNN & UNet + attention gates & -- & full \\
SegResNet & CNN & Residual encoder-decoder & -- & full \\
FlexibleUNet & CNN & EfficientNet-B0 encoder + UNet decoder & -- & full \\
Swin UNETR & Transformer & Swin encoder + CNN decoder & -- & full \\
SAM (ViT-B) & Foundation model & ViT-B encoder + SAM decoder + adapters & SA-1B & adapters only \\
MedSAM (ViT-B) & Foundation model & ViT-B encoder + SAM decoder + adapters & 1.5M medical (incl. CT) & adapters only \\
\bottomrule
\end{tabular}%
}
\end{table}

We compare LegSegNet with a broad set of 2D segmentation baselines, including convolution-based and transformer-based architectures. Specifically, we train five models from scratch: UNet \cite{ronneberger2015u}, Attention UNet \cite{oktay2018attention}, SegResNet \cite{myronenko20183d}, FlexibleUNet with an EfficientNet-B0 encoder \cite{tan2019rethinking}, and SwinUNETR \cite{hatamizadeh2021swin}. In addition, we evaluate finetuned ViT-B versions of SAM \cite{kirillov2023segment} and MedSAM \cite{ma2024segment} for automated multi-class segmentation. Following the adapter-based finetuning strategies, only the inserted adapter layers are trainable, while the original foundation model parameters are frozen.

\subsection{Evaluation Metrics}

We evaluate segmentation performance using Dice similarity coefficient (DSC) and average symmetric surface distance (ASSD). DSC measures the overlap between the predicted mask $P$ and the ground-truth mask $G$:

\begin{equation}
\mathrm{DSC}(P, G) = \frac{2|P \cap G|}{|P| + |G|}
\end{equation}

A higher DSC indicates better model performance. ASSD measures the average boundary distance between the predicted surface $S_P$ and the ground-truth surface $S_G$:

\begin{equation}
\mathrm{ASSD}(S_P, S_G) = \frac{\sum_{p \in S_P} d(p, S_G) + \sum_{g \in S_G} d(g, S_P)}{|S_P| + |S_G|}
\end{equation}

A lower ASSD indicates better boundary accuracy. Both metrics are computed separately for bone, skeletal muscle, subcutaneous adipose tissue, and inter/intramuscular adipose tissue at volume level.

\subsection{Implementation Details}

LegSegNet is implemented using nnUNet v2 \cite{isensee2021nnu}. We evaluate three 2D nnUNet configurations: the default plain UNet, ResEnc-L, and ResEnc-XL. All three models use the default nnUNet training pipeline without modification, including automatic preprocessing, patch size selection, data augmentation, deep supervision, SGD optimization, polynomial learning-rate decay, and training for 1000 epochs. The automatically selected patch size is 192 $\times$ 256.

\begin{figure*}[t]
\centering
\includegraphics[width=0.95\linewidth]{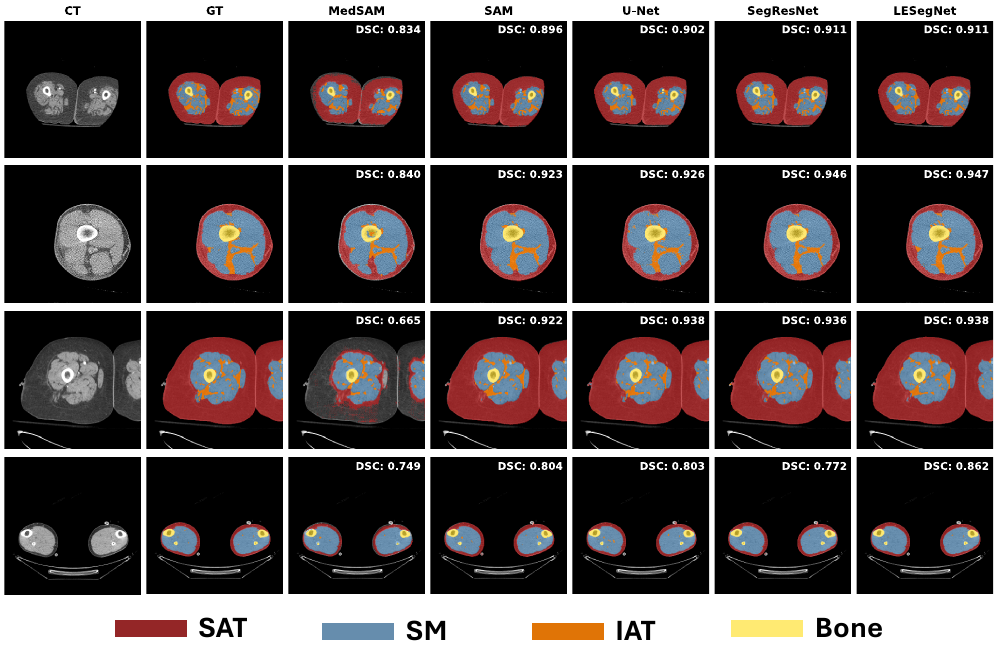}
\caption{\textbf{Qualitative segmentation results.} Example CT slices from the test set with ground truth annotations and predictions from LegSegNet. Colors indicate different tissue types: bone (yellow), skeletal muscle (blue), subcutaneous adipose tissue (red), and inter/intramuscular adipose tissue (orange).}
\label{fig:results}
\end{figure*}

For comparison, we train additional 2D baselines from MONAI, including UNet, Attention UNet, SegResNet, FlexibleUNet with an EfficientNet-B0 encoder, and SwinUNETR. For preprocessing, CT intensities (HU) are clipped to $[-200, 200]$ and normalized to $[0, 255]$. These models are trained from scratch for 1000 epochs with input resolution of 256 $\times$ 256, AdamW optimization, and standard spatial and intensity augmentations. We also finetune SAM and MedSAM in an automated multi-class setting for up to 300 epochs using adapters following \cite{gu2025segmentanybone}, while keeping the original foundation model weights frozen. All experiments are run using a single NVIDIA A6000 GPU.

\begin{figure*}[t]
\centering
\includegraphics[width=1\linewidth]{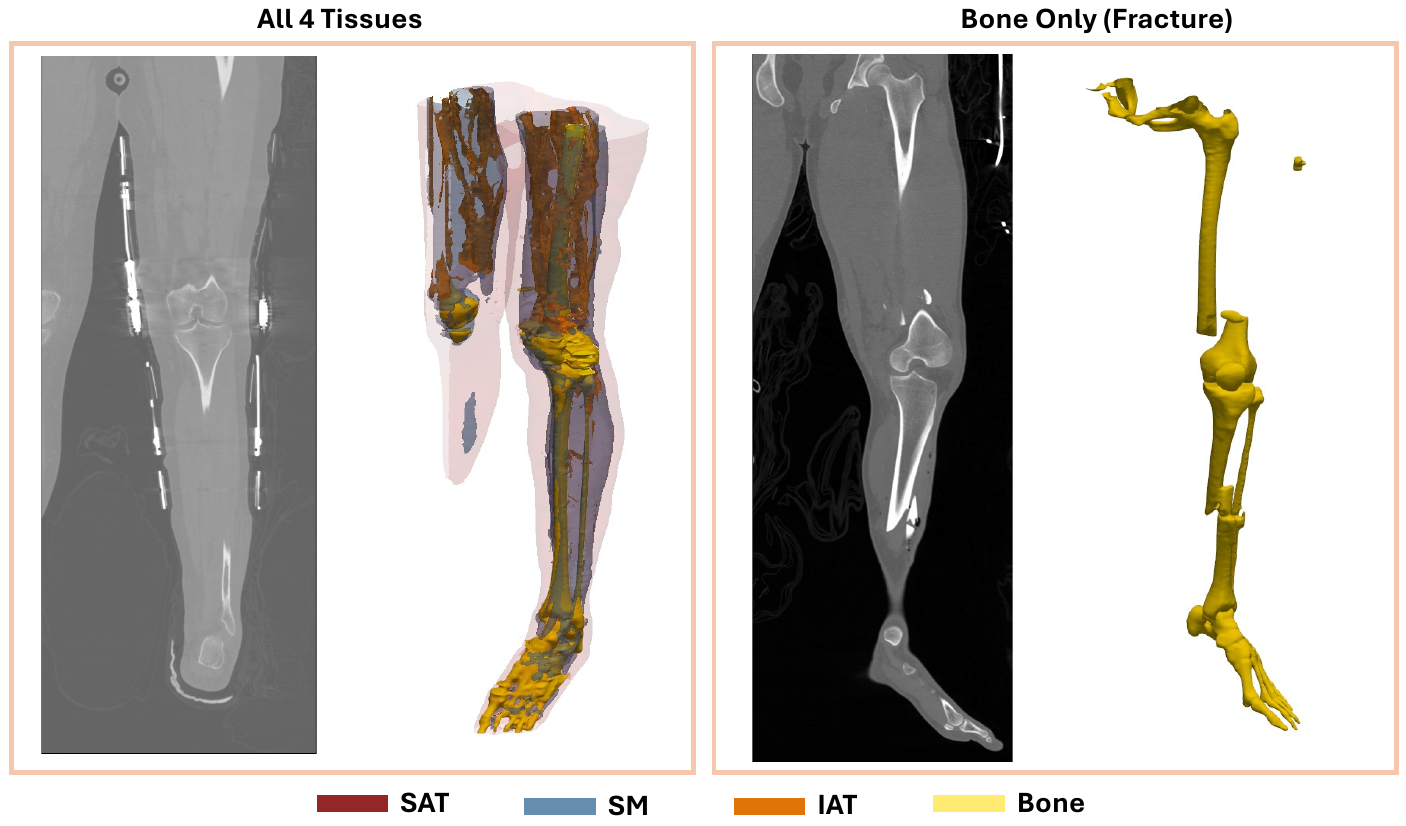}
\caption{\textbf{3D visualization of predicted segmentation.} Volumetric reconstruction of LegSegNet predictions.}
\label{fig:3d_view}
\end{figure*}

\begin{figure*}[t]
\centering
\includegraphics[width=\textwidth]{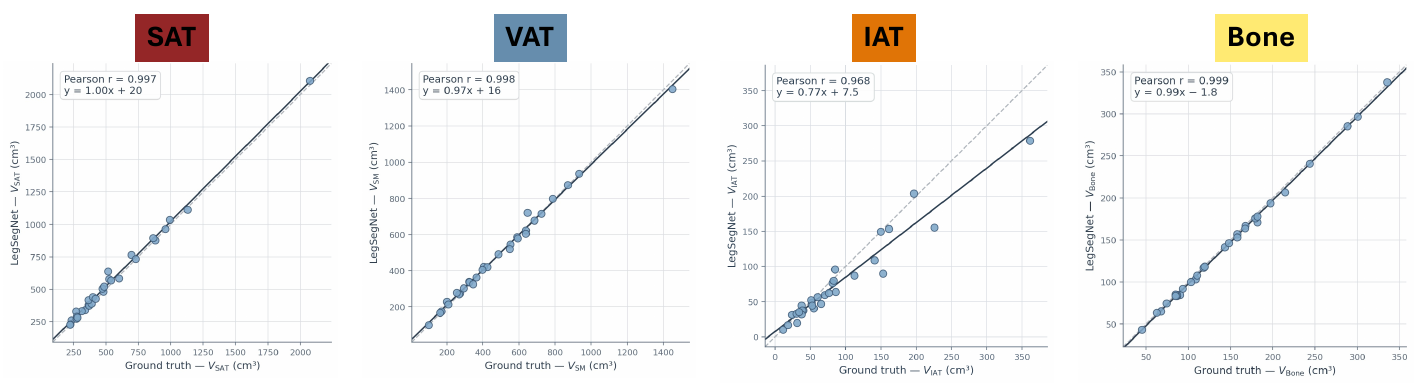}
\caption{\textbf{Correlation between predicted and ground-truth tissue volumes.} Pearson correlation between tissue volumes computed from LegSegNet and ground truth volumes for \textbf{SAT}, \textbf{SM}, \textbf{IAT}, and \textbf{Bone} on the held-out test set.}
\label{fig:correlation}
\end{figure*}

\subsection{Results}

\subsubsection{Segmentation}
Table \ref{tab:test_results_3d} presents the quantitative segmentation performance across all methods. The default nnUNet (2D) achieves the highest average DSC of 89.31\% and the lowest average ASSD of 0.515. The residual encoder variants (ResEnc-L and ResEnc-XL) achieve comparable overall performance, with ResEnc-L obtaining the best SAT DSC (93.27\%).

Among CNN-based methods, SegResNet performs best with an average DSC of 88.56\%, followed by UNet and Attention UNet. For Transformer-based methods, SwinUNETR achieves 86.96\% average DSC. Finetuned SAM and MedSAM obtain average DSC values of 86.40\% and 79.17\%, respectively.

Across all tissues, bone segmentation achieves the highest performance, with Dice scores exceeding 95\% for all methods except MedSAM. IAT segmentation achieves the lowest performance, with the best DSC of 73.98\% obtained by the default nnUNet. The qualitative results in Figure \ref{fig:results} show that LegSegNet produces anatomically consistent masks across representative axial slices, and the 3D reconstruction in Figure \ref{fig:3d_view} further illustrates the volumetric continuity of the predicted lower extremity tissue segmentation.

In addition, LegSegNet achieves a comparable \textbf{SAT} performance on the external test (92.13\% DSC). However, the DSC of \textbf{SM} seems to decrease to around 87\% due to different annotation protocols, as shown in Figure \ref{fig:results_saros}.

\begin{table}[t]
\centering
\caption{Model performance on SAROS dataset including per-class 3D Dice ($\uparrow$) with 95\% bootstrap confidence intervals. The best is bold and runner-up is underlined.}
\label{tab:test_results_saros_3d}
\resizebox{\linewidth}{!}{
\begin{tabular}{lccc}
\toprule
Method & \textbf{SAT} & \textbf{SM} & Ave. \\
\midrule
\multicolumn{4}{l}{\textit{LegSegNet}} \\
  nnUNet (default 2D) & \textbf{92.13 [91.13, 93.06]} & 86.62 [85.53, 87.77] & 89.38 [88.61, 90.16] \\
  nnUNet ResEnc-L & \underline{92.04 [91.00, 93.04]} & \underline{86.87 [85.85, 87.97]} & \underline{89.45 [88.71, 90.19]} \\
  nnUNet ResEnc-XL & 91.94 [90.86, 92.98] & \textbf{87.02 [86.03, 88.09]} & \textbf{89.48 [88.72, 90.26]} \\
\bottomrule
\end{tabular}
}
\end{table}

\subsubsection{Body Composition Correlation}
To evaluate whether segmentation performance translates into reliable body composition measurements, we also compare LegSegNet-derived tissue volumes with reference volumes computed from the ground truth masks. This analysis is performed on stacked 3D test volumes. As shown in Figure \ref{fig:correlation}, the Pearson correlation coefficients are 0.997 for \textbf{SAT}, 0.998 for \textbf{SM}, 0.968 for \textbf{IAT}, and 0.999 for \textbf{Bone}.

\begin{figure}[H]
\centering
\includegraphics[width=\linewidth]{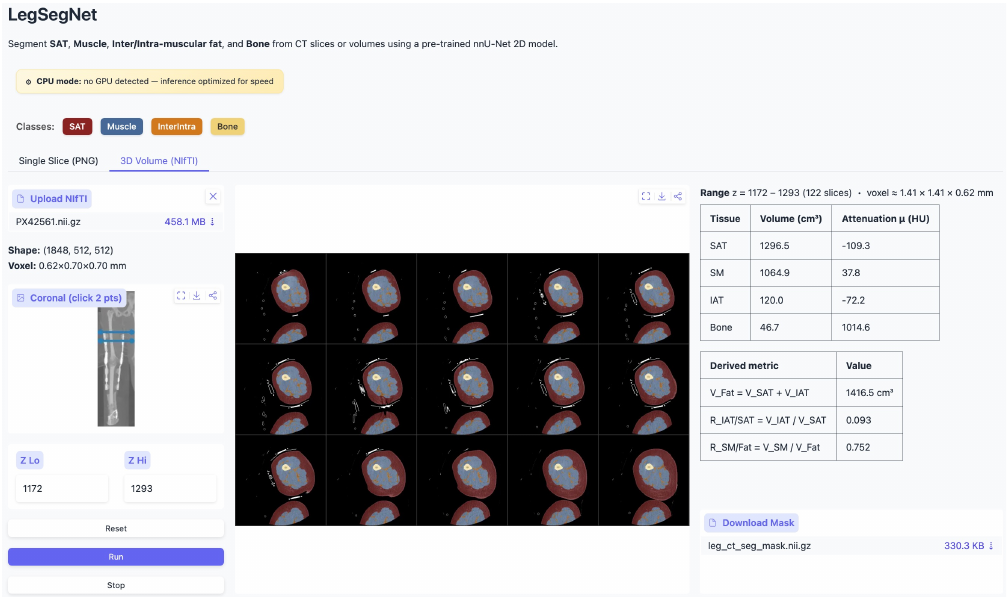}
\caption{\textbf{Overview of LegSegNet system.} LegSegNet takes lower extremity CT scans as input and outputs tissue masks, visualizations, and body composition measurements for downstream analysis.}
\label{fig:system}
\end{figure}

\subsection{LegSegNet System}
As shown in Figure \ref{fig:pipeline} and Figure \ref{fig:system}, LegSegNet provides an end-to-end workflow for lower extremity CT analysis. Given CT scans, the user can specify the axial range for analysis, after which the system automatically starts segmentation within the selected region. The predicted masks are then used to generate visual outputs including slice-level overlays, as well as quantitative body composition measurements such as tissue volume, attenuation, and ratios between different tissue volume.

\subsection{Discussion}
\paragraph{Performance analysis.} Our benchmark demonstrates that LegSegNet, based on nnUNet, consistently outperforms CNN-based, transformer-based, and finetuned foundation-model baselines. The default 2D nnUNet achieves the best overall performance with an average DSC of 89.31 and ASSD of 0.515. For class-wise results, ResEnc-L obtains the highest \textbf{SAT} DSC (93.27), while the default 2D nnUNet achieves the best performance on \textbf{SM}, \textbf{IAT}, and \textbf{Bone}. These results suggest that nnUNet's self-configuring pipeline effectively adapts to lower extremity CT characteristics without manual tuning. The residual encoder variants do not improve overall performance over the default 2D nnUNet, indicating that increasing model capacity alone may not add further benefits to this task.

\begin{figure}[H]
\centering
\includegraphics[width=\linewidth]{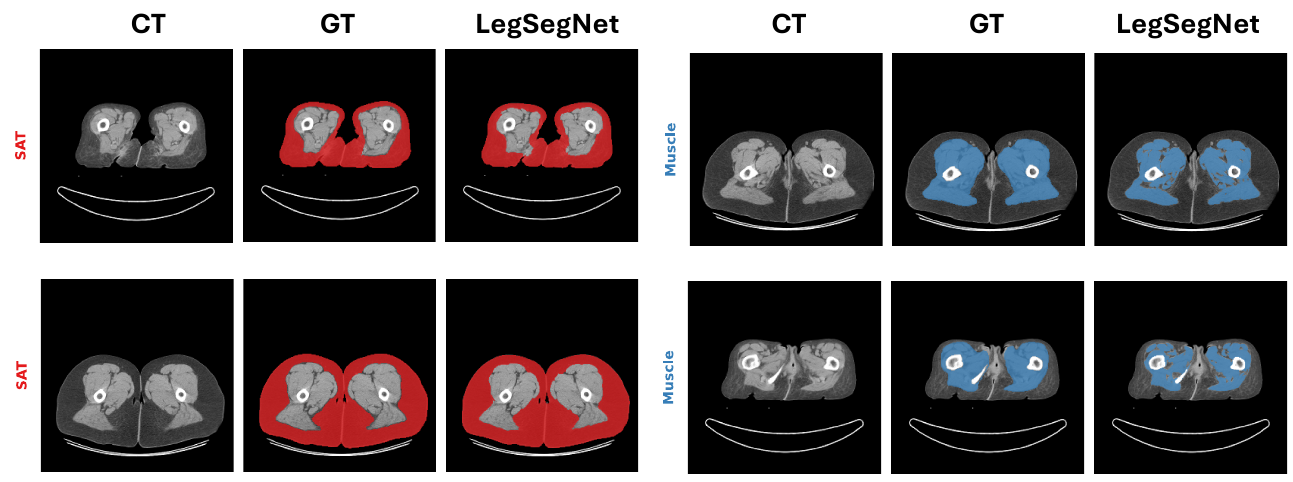}
\caption{\textbf{Qualitative SAROS segmentation results.} Example SAROS CT slices with ground truth annotations (\textbf{SAT} and \textbf{SM}) and predictions from LegSegNet.}
\label{fig:results_saros}
\end{figure}

The performance drop for \textbf{SM} on SAROS is mainly due to differences in annotation protocols. As shown in Figure \ref{fig:results_saros}, \textbf{IAT} is included within \textbf{SM} in the SAROS masks, whereas our protocol labels \textbf{IAT} as a separate class. Moreover, LegSegNet maintains stable performance for \textbf{SAT} on SAROS, demonstrating good generalizability of our model.

\paragraph{Class-wise performance.} Bone segmentation is the most reliable target, likely because cortical bone has strong CT contrast and well-defined boundaries. In contrast, \textbf{IAT} remains the most challenging class because it is relatively sparse and can have similar intensity to adjacent adipose tissue compartments (e.g., \textbf{SAT}). This suggests that future improvements should focus on more consistent \textbf{IAT} annotation, and including more representative cases with diverse fat distribution.

\paragraph{Body composition analysis.} The strong tissue-volume correlations in Figure \ref{fig:correlation} indicate that high segmentation accuracy also translates into reliable quantitative measurements. The correlations are especially high for \textbf{SAT}, \textbf{SM}, and \textbf{Bone}, while \textbf{IAT} remains slightly lower, consistent with its lower segmentation DSC. These results support the use of LegSegNet not only for mask generation but also for automated lower extremity CT body composition analysis.

\section{Conclusion}
\label{sec:conclusion}

In this work, we present LegSegNet, an end-to-end deep learning system for lower extremity CT tissue segmentation and body composition quantification. LegSegNet can segment bone, skeletal muscle, SAT, and IAT, and convert the predicted masks into quantitative measurements including tissue volume, attenuation, and tissue volume ratios. Our comprehensive benchmark demonstrates that LegSegNet achieves the best overall segmentation performance compared with CNN-based, transformer-based, and finetuned foundation model-based baselines. In addition, the strong correlation between predicted and ground truth tissue volumes shows that LegSegNet preserves clinically relevant quantitative measurements for downstream body composition analysis. By releasing the system and model weights, we believe LegSegNet can serve as a public system for evaluating future computer vision models on clinically meaningful CT tissue segmentation and quantification tasks.

{
    \small
    \bibliographystyle{ieeenat_fullname}
    \bibliography{reference}
}

\end{document}